\documentclass[conference]{IEEEtran}
\usepackage{times}
\usepackage{cite}
\usepackage[T1]{fontenc} 
\usepackage{textcomp}   
\usepackage{amsmath, amssymb}
\usepackage{graphicx}
\usepackage{subcaption}
\usepackage{mathtools}
\usepackage{algorithm}
\usepackage{algorithmicx}
\usepackage{algpseudocode}
\usepackage{booktabs} 
\usepackage{url}
\usepackage{pifont}
\usepackage{xcolor}
\usepackage{footmisc}
\usepackage{multirow}
\usepackage{rotating}
\usepackage{mathrsfs}

\usepackage{multicol}
\usepackage[bookmarks=true]{hyperref}

\usepackage{amsthm}

\title{Function-Space Diffusion for Motion Planning}

\IEEEoverridecommandlockouts
\author{Zinuo Chang, Yipu Chen, Byoungwoo Park, Hongzhe Yu, and Yongxin Chen
\thanks{Z. Chang is with the School of Electrical and Computer Engineering, Yipu Chen, H. Yu and Yongxin Chen are with the School of Aerospace Engineering, Georgia Institute of Technology, Atlanta, GA; {\{zchang40, ychen3302, hyu419, yongchen\}@gatech.edu}}
\thanks{B. Park is with Korea Advanced Institute of Science and Technology, Daejeon, Republic of Korea; {bw.park@kaist.ac.kr}}
}

\begin{document}

\maketitle

\begin{abstract}
Diffusion-based motion planners have demonstrated strong performance in generating diverse and high-quality robot trajectories in cluttered environments with multiple feasible solutions. However, existing approaches typically operate on fixed-length waypoint sequences, making the learned model resolution-dependent, thereby preventing zero-shot generalization across resolutions.
In this work, we propose Function-Space Diffusion for Motion Planning (FSD-MP), a diffusion-based motion planner that models trajectories as continuous functions and performs diffusion directly in function space, achieving discretization-invariant trajectory generation.
We define a mode-wise forward process in the spectral domain, driven by Gaussian noise with a Mat\'ern-type covariance, and parameterize the reverse process with a boundary-compatible Discrete Sine Transform-based Fourier Neural Operator (DST-FNO) that preserves start-goal constraints across resolutions.
We evaluate FSD-MP on 2D point robot and 7-DoF Franka manipulator planning benchmarks. Our method achieves competitive planning performance at the training resolution and generalizes zero-shot across resolutions up to 16× higher, preserving consistent planning behavior without retraining. These results demonstrate that function-space diffusion provides an effective framework for discretization-invariant motion planning.
\end{abstract}

\section{Introduction}
\label{sec:introduction}
Motion planning is a crucial component of robotic systems, aiming to find smooth and collision-free trajectories that connect start and goal configurations in cluttered environments. The problem is inherently non-convex due to obstacles, and often admits multiple distinct solutions.

Classical motion planning methods have been extensively studied from both optimization-based and sampling-based perspectives. Trajectory optimization methods refine an initial trajectory by minimizing costs related to smoothness, collision avoidance, and task objectives~\cite{ratliff2009chomp, kalakrishnan2011stomp, mukadam2018continuous, yu2023gaussian}. While often effective in practice, they are sensitive to initialization and may converge to poor local minima or infeasible solutions under non-convex collision constraints. Sampling-based planners explore the configuration space and provide probabilistic completeness guarantees~\cite{kavraki1996probabilistic,karaman2011sampling, kuffner2000rrt}, but their exploration can become inefficient in high-dimensional spaces or narrow passages. Moreover, the resulting paths often require post-processing to satisfy smoothness or control-related requirements.

These limitations have motivated learning-based motion planners that directly generate feasible trajectories by learning from expert demonstrations. By capturing the distribution of feasible trajectories, such models can rapidly generate diverse, high-quality solutions without exhaustive online search or carefully designed initialization. Among these approaches, diffusion-based generative models~\cite{janner2022diffuser, carvalho2023mpd, carvalho2025mpd} have proven to be a powerful framework for modeling complex, multimodal trajectory distributions, making them well-suited for motion planning in cluttered and high-dimensional spaces.

Despite their promise, existing diffusion-based planners represent trajectories as fixed-length waypoint sequences. This tightly couples the learned model to a specific discretization: a model trained at one resolution cannot generalize to another without retraining. This dependence is undesirable in motion planning, where trajectories are naturally continuous, and the appropriate resolution may vary across tasks, environments, and downstream controllers. Coarse discretizations are computationally efficient but may miss geometric details near obstacles, whereas fine discretizations improve fidelity at higher computational cost.

To address this gap, we propose \textit{Function-Space Diffusion for Motion Planning} (FSD-MP), which achieves discretization invariance by treating trajectories as continuous functions and performing diffusion directly in function space. This formulation enables zero-shot generalization to arbitrary resolutions without retraining. We demonstrate that a model trained at a base resolution can generate trajectories at up to 16× higher resolution while converging to the same solution modes, maintaining both feasibility and multimodality.

Our key contributions are:

\begin{itemize}
    \item A function-space diffusion framework for motion planning that is inherently discretization-invariant, enabling flexible trajectory generation at arbitrary resolutions without retraining.
    \item A boundary-compatible neural operator architecture that consistently satisfies start-goal constraints across resolutions.
    \item Experimental demonstration of zero-shot super-resolution up to 16× the training resolution, while preserving consistent planning behavior across resolutions.
\end{itemize}

\begin{figure*}[htpb]
    \centering
    \includegraphics[width=0.95\linewidth]{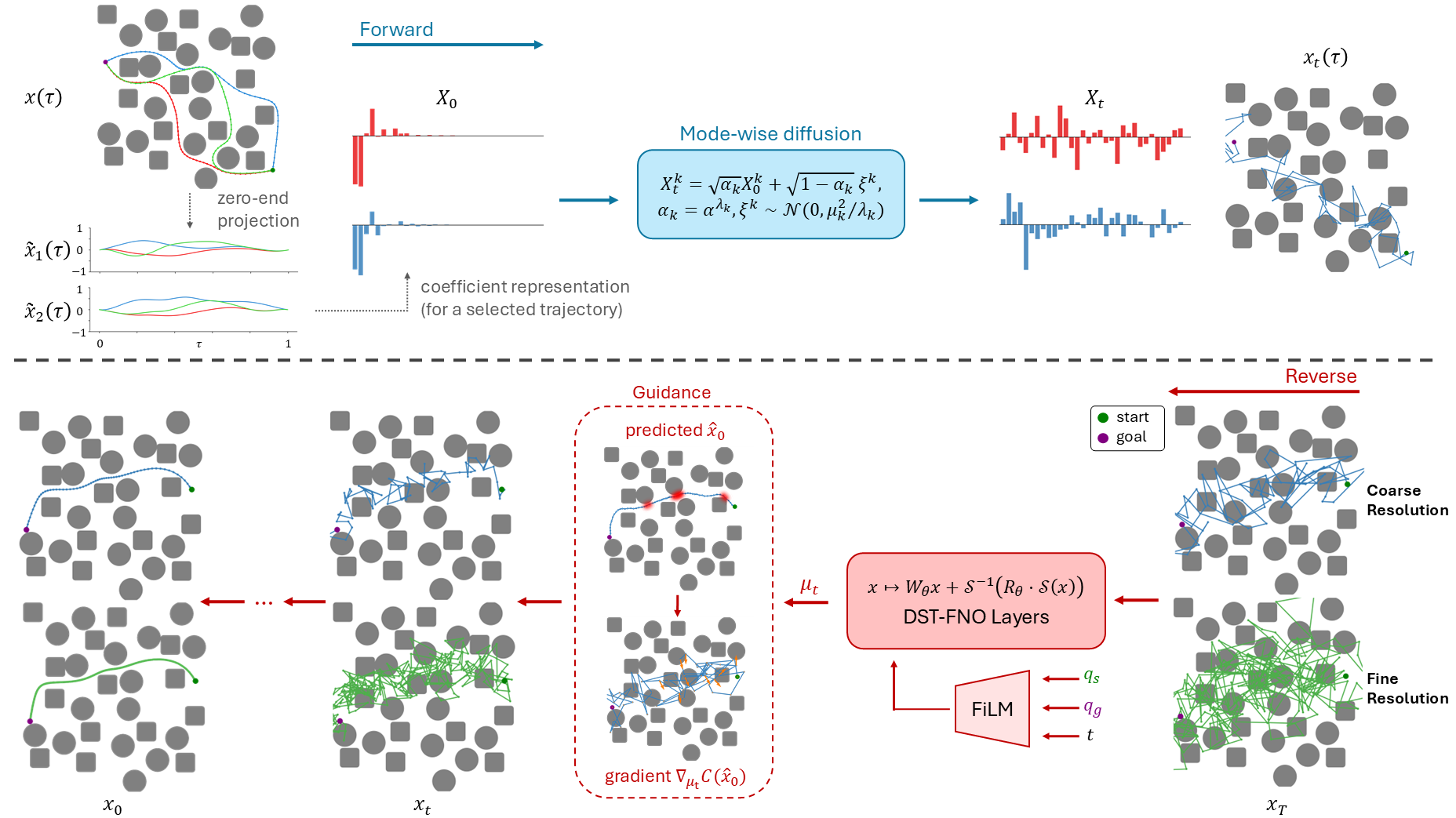}
    \caption{Overview of FSD-MP. \textbf{(Top)} The forward process performs zero-end projection, transforms trajectories into DST coefficients and applies mode-wise diffusion with variance scaled by eigenvalue $\lambda_k$. \textbf{(Bottom)} The reverse process denoises via FiLM-conditioned DST-FNO layers, together with guidance based on the predicted clean trajectory $\hat{x}_0$. The function-space formulation enables \textbf{discretization-invariant} trajectory generation across both coarse and fine discretizations.
    }
\end{figure*}

\section{Related Work}
\label{sec:related_work}

\paragraph{Learning-based Planning}
Learning-based methods aim to encode trajectory distributions from demonstrations or expert planners, providing informative biases for sampling, optimization, or direct trajectory generation. Early approaches use Gaussian Mixture Models and Probabilistic Movement Primitives~\cite{calinon2007learning, paraschos2013probabilistic}, but such models scale poorly to high-dimensional configuration spaces.
More recent approaches adopt deep generative models, including generative adversarial networks (GANs), variational autoencoders (VAEs), normalizing flows, and energy-based models (EBMs), to learn richer trajectory distributions for planning~\cite{lembono2021learning, ichter2018learning, power2024learning, urain2022learning}. However, these models often require trade-offs among distributional expressiveness, training stability, and conditional generation flexibility, motivating diffusion models as a flexible framework for modeling complex, multimodal trajectory distributions.

\paragraph{Diffusion Models in Robotics}
Diffusion models generate samples by learning to reverse an iterative noising process~\cite{ho2020ddpm, song2020score}. Their expressive modeling capacity and flexible conditioning have led to broad adoption in robotics, including policy learning, manipulation, and decision-making~\cite{chi2025diffusion, mishra2023generative, ajay2023conditional,mishra2024reorientdiff}.
In motion planning, Diffuser~\cite{janner2022diffuser} pioneered the application of diffusion models to offline reinforcement learning and trajectory planning. Motion Planning Diffusion (MPD)~\cite{carvalho2023mpd} further adapted diffusion for robot manipulation to generate diverse, collision-free trajectories conditioned on task goals.
These works demonstrate the effectiveness of diffusion models for capturing multimodal trajectory distributions. However, most existing diffusion-based methods operate on finite-dimensional waypoint or action sequences with a fixed discretization, limiting their ability to generalize across different resolutions.

\paragraph{Diffusion Models in Function Space}
Recent works have extended diffusion and score-based generative models to infinite-dimensional function spaces~\cite{lim2025score, kerrigan2023diffusion, pidstrigach2024infinite}, enabling discretization-invariant generation. These function-space formulations have been applied to diverse domains with naturally continuous structure, including image generation and physics-informed problems governed by PDEs such as Navier--Stokes.
Despite this progress, function-space diffusion remains largely unexplored in robotics and motion planning, even though trajectories are inherently continuous functions and practical deployment often requires flexible discretizations. Our work addresses this gap by formulating motion planning as diffusion over trajectory functions, enabling zero-shot generalization across arbitrary discretizations.

\section{Problem Formulation}
\label{sec:formulation}
Motion planning involves finding a smooth, collision-free trajectory between the start and goal configurations $q_s, q_g \in \mathbb{R}^d$, where $d$ denotes the dimension of the robot configuration space. We represent a trajectory as a continuous function
\begin{equation}
    x(\tau):[0,1]\rightarrow \mathbb{R}^d
\end{equation}
where $\tau \in [0, 1]$ denotes normalized time.
The planning problem is to find an optimal trajectory $x^\star(\tau)$ that minimizes a cost functional $J$ for a given set of end-point configurations:
\begin{equation}
    x^\star = \arg\min_{x}\ J(x)
    \quad \text{s.t.}\quad x(0)=q_s,\ x(1)=q_g,
\end{equation}
with the objective decomposed as
\begin{equation}
    J(x) = J_\mathrm{prior}(x) + J_\mathrm{task}(x),
\end{equation}
where $J_\mathrm{prior}$ encourages smoothness or consistency with prior knowledge of feasible trajectories, and $J_\mathrm{task}$ penalizes task-dependent constraint violations in the environment $\mathcal{O}$.

We adopt a probabilistic perspective and view motion planning as posterior inference over trajectory functions: 
\begin{equation}
    p(x \mid \mathcal{O}) \propto  p(x)\ p(\mathcal{O}\mid x)
\end{equation}
where $p(x)$ represents a prior distribution over smooth, feasible trajectories, and the likelihood $p(\mathcal{O}\mid x)$ encodes task-specific constraints such as collision avoidance and joint limits. 

This posterior perspective generalizes deterministic optimization by capturing uncertainty and multimodality in feasible motions. Our goal is to generate collision-free trajectories by sampling from the posterior $p(x\mid\mathcal{O})$.

\section{Function-Space Diffusion Motion Planning}
\label{sec:methods}
We propose \textit{Function-Space Diffusion for Motion Planning} (FSD-MP), which samples from the posterior $p(x|\mathcal{O})$ defined in Section~\ref{sec:formulation} while achieving discretization invariance. 
Rather than parameterizing trajectories as fixed-length waypoint sequences, FSD-MP represents each trajectory as a continuous function $x(\tau) \in \mathcal{H}$, in a Hilbert space $\mathcal{H} = L^2([0,1], \mathbb{R}^d)$. The model learns the function-space prior $p(x)$, while the likelihood $p(\mathcal{O}|x)$ is incorporated at inference time via classifier guidance~\cite{dhariwal2021diffusion,carvalho2023mpd}. This function-space formulation enables trajectory generation at arbitrary resolutions without retraining, as discretization is introduced only during evaluation.

\subsection{Function-Space Forward Diffusion}
We define the forward diffusion process over trajectories as a linear SDE in the trajectory space $\mathcal{H}$:
\begin{equation}
    \mathrm{d} x_t=-\frac{1}{2} \, \beta(t) \, \mathcal{F} x_t \mathrm{d} t + \sqrt{\beta(t)}\, \mathcal{G}\mathrm{d} W_t
    \label{eq:inf-sde}
\end{equation}
where $x_t \in \mathcal{H}$ is the noisy trajectory function at diffusion time $t\in [0, T]$, $\beta(t) \geq 0$ is a noise schedule, $\mathcal{F}: \mathcal{D}(\mathcal{F}) \subset \mathcal{H} \to \mathcal{H}$ is the linear drift operator, $\mathcal{G}: \mathcal{H} \to \mathcal{H}$ is the linear diffusion operator, and $W_t$ is a cylindrical Wiener process. The corresponding noise covariance operator is $\mathcal C=\mathcal G\mathcal G^\ast$.

This construction generalizes the Variance Preserving (VP) SDE to function space: $\mathcal{F}$ controls the diffusion rate across spectral modes, while $\mathcal{G}$ determines the covariance geometry of the injected noise.

To make~\eqref{eq:inf-sde} analytically tractable, we assume $\mathcal{F}$ and $\mathcal{G}$ share an orthonormal eigenbasis $\{\phi_k\}_{k\in\mathbb{N}}$ with eigenvalues
\begin{equation*}
    \mathcal{F} \phi_k = \lambda_k \phi_k, \quad \mathcal{G} \phi_k = \mu_k \phi_k, \quad \lambda_k, \mu_k >0
\end{equation*}

Projecting $x_t$ onto each spectral mode, $X_t^k = \langle x_t, \phi_k \rangle$, the SDE~\eqref{eq:inf-sde} decouples into independent coefficient SDEs:
\begin{equation}
    \mathrm{d} X^k_t = -\frac{1}{2}\lambda_k \beta(t)X^k_t \mathrm{d} t+\mu_k \sqrt{\beta(t)}\mathrm{d} W_t^k, \quad k\in\mathbb{N}
    \label{eq:inf-sde-freq}
\end{equation}
where $\{W_t^k\}$ are independent standard Brownian motions. This modal decomposition admits closed-form marginals:
\begin{equation}
    X_t^k = \sqrt{\alpha_k(t)}X_0^k + \sigma_k(t) \epsilon^k, \quad \epsilon^k \sim \mathcal{N}(0, 1),
    \label{eq:closed-form-in-projection}
\end{equation}
\begin{equation*}
\begin{aligned}
    \text{where} \quad \alpha_k(t) &= \exp\Big(-\lambda_k\int_0^t \beta(s)\mathrm{d}s\Big) = \alpha(t)^{\lambda_k} \\
    \sigma_k^2(t) &= \frac{\mu_k^2}{\lambda_k}(1 - \alpha_k(t))
\end{aligned}
\end{equation*}
and $\alpha(t)$ is the noise schedule in standard diffusion.

To simplify the expression, we introduce a rescaled noise variable $\xi\sim \mathcal{N}(0, \mathcal{F}^{-1}\,\mathcal{C})$, whose coefficients satisfy
\begin{equation}
     \xi^k := \langle \xi,\phi_k\rangle =
    \frac{\mu_k}{\sqrt{\lambda_k}}\epsilon^k
\end{equation}

With this reparameterization, the forward process in~\eqref{eq:closed-form-in-projection} admits a compact form analogous to standard diffusion:
\begin{equation}
    X_t^k = \sqrt{\alpha_k(t)}X_0^k + \sqrt{1-\alpha_k(t)}\,\xi^k.
    \label{eq:closed-form-xi}
\end{equation}

This yields a training objective that decomposes over modes (see Appendix~\ref{sec:objective}):
\begin{equation}
    \mathcal L(\theta) = \mathbb E_{t,x_0,\xi}
    \left[
        \sum_{k=1}^{\infty}
        \left|
            \xi_\theta^k(x_t,t)-\xi^k
        \right|^2
    \right].
    \label{eq:xi-loss-modes}
\end{equation}

By Parseval's identity, this is equivalent to minimizing the MSE of $\xi$ in the trajectory domain:
\begin{equation}
    \mathcal L(\theta)
    =
    \mathbb E_{t,x_0,\xi}
    \bigl[
        \|\xi_\theta(x_t,t)-\xi\|_{\mathcal H}^2
    \bigr].
    \label{eq:xi-loss-time-domain}
\end{equation}

\subsubsection*{Well-Defined Noise in Function Spaces}
A critical challenge in infinite-dimensional diffusion is defining valid Gaussian noise in $\mathcal{H}$. A Gaussian measure $\mathcal{N}(0, \mathcal{C})$ is supported on $\mathcal{H}$ only if its covariance $\mathcal{C}$ is trace-class~\cite{lim2025score, pidstrigach2024infinite}. This rules out white noise, whose identity covariance $\mathcal{I}$ is not trace-class.
We therefore adopt a Mat\'ern-type covariance operator:
\begin{equation}
    \mathcal C=\sigma^2\left(-\Delta+\kappa^2  \mathcal I\right)^{-\alpha},
\end{equation}
where $\Delta$ is the Laplacian under the chosen boundary condition, and $\sigma, \kappa, \alpha > 0$ control variance, length scale, and smoothness. For time-parameterized trajectories, $\mathcal C$ is trace-class when $\alpha > 1/2$. Details are provided in Appendix~\ref{sec:noise}.

\subsection{Resolution-Agnostic Neural Operator}
\label{sec:fno}
The reverse diffusion process requires learning a noise prediction operator $\xi_\theta(x_t, t): \mathcal{H} \times [0, T] \to \mathcal{H}$.
Standard diffusion models typically parameterize this mapping with architectures like U-Net \cite{ronneberger2015u, ho2020ddpm} operating on fixed-size vectors in $\mathbb{R}^{N \times d}$, making them inherently resolution-dependent.

To achieve discretization invariance, we parameterize $\xi_\theta$ as a neural operator that learns the mapping between function spaces. We adopt Fourier Neural Operator (FNO)~\cite{li2021fourier}, which operates in the Fourier domain by learning spectral kernels that act on frequency modes rather than spatial discretizations.

An FNO model consists of a lifting layer $P$ that embeds the input into feature space, a series of Fourier layers, and a projection layer $Q$ mapping back to the trajectory space. Each \emph{Fourier layer} defines an operator $\mathcal{N}$ that transforms a feature function $u:[0,1]\to\mathbb{R}^c$, where $c$ denotes the hidden channel dimension, as:
\begin{equation}
    (\mathcal{N} u)(\tau) = \sigma\left(Wu(\tau) + \mathscr{F}^{-1}(R \cdot \mathscr{F}u)(\tau)\right)
    \label{eq:fno-layer}
\end{equation}
where $\mathscr{F}$ and $\mathscr{F}^{-1}$ denote the Fast Fourier Transform (FFT) and its inverse, $R\in \mathbb{C}^{K \times c \times c}$ contains learned complex weights applied to the first $K$ modes, $W \in \mathbb{R}^{c \times c}$ is a pointwise linear transform, and $\sigma$ is a nonlinear activation.

This frequency-domain transformation enables the same operator to be evaluated at different resolutions without retraining.

\subsection{Endpoint Constraint Enforcement}
In motion planning problems, trajectories must satisfy fixed endpoint constraints: $x(0)=q_s$ and $x(1)=q_g$. 
Existing diffusion planners such as Diffuser and MPD~\cite{janner2022diffuser, carvalho2023mpd} enforce this through inpainting, directly replacing endpoint values during training and sampling. However, in our function-space formulation, the trajectories are continuous functions with global spatial correlation. Naively fixing endpoints will therefore introduce inconsistencies and degrade sampling stability.
We address this through three components: zero-end projection, endpoint-conditioning via FiLM, and a boundary-compatible DST-based neural operator.

\paragraph{\textbf{Zero-End Projection}}
We enforce endpoint constraints by reformulating constrained trajectory generation 
as a zero-boundary residual learning problem:
\begin{equation}
    \tilde{x}(\tau) = x(\tau) - (1-\tau)q_{\mathrm{s}} - \tau q_{\mathrm{g}}
    \label{eq:zero-end-projection}
\end{equation}
which satisfies $\tilde{x}(0) = \tilde{x}(1) = 0$ by construction. We apply this projection to all trajectories during training, so the model learns the distribution over zero-boundary residuals $p(\tilde{x})$. 

At inference, we sample a residual function $\tilde{x}(\tau)$ and recover the full trajectory via
\begin{equation}
    x(\tau) = \tilde{x}(\tau) + (1-\tau)q_{\mathrm{s}} + \tau q_{\mathrm{g}}
    \label{eq:reconstruction}
\end{equation}

Since \eqref{eq:reconstruction} is a deterministic affine transformation applied after sampling, the reconstructed trajectory is \emph{guaranteed} to satisfy $x(0) = q_{\mathrm{s}}$ and $x(1) = q_{\mathrm{g}}$ exactly, without post-hoc endpoint clamping or iterative correction.

\paragraph{\textbf{Conditional Generation via FiLM}}
The zero-end projection removes absolute position information from trajectories, making the residual distribution $p(\tilde{x})$ endpoint-dependent. We therefore condition the neural operator on the endpoint information using Feature-wise Linear Modulation (FiLM)~\cite{perez2018film}, which modulates intermediate feature functions $u$ via learned affine transformations:
\begin{equation}
    \text{FiLM}(u; \gamma, \delta) = \gamma \odot u + \delta
    \label{eq:film}
\end{equation}
where $\gamma, \delta \in \mathbb{R}^{c}$ are channel-wise scale and shift parameters. These modulation parameters are generated from the conditioning vector $\mathbf{c} = [q_s; q_g; \mathrm{emb}(t)]$ through a per-layer MLP:
\begin{equation}
    \gamma_\ell, \delta_\ell = \text{MLP}_\ell(\mathbf{c})
\end{equation}
where $\ell$ indexes the neural-operator layer. This allows the network to adapt its predictions to each start-goal configuration while preserving the zero-boundary residual representation.

\begin{algorithm}[t]
\caption{Training with Zero-End Projection}
\label{alg:train}
\begin{algorithmic}[1]
\Require Collision-free trajectories $\mathcal{D}$, diffusion model $\xi_\theta$, spectral schedule $\alpha_k(t)$, learning rate $\eta$
\While{training not finished}
    \State Sample $x_{0} \sim \mathcal{D}$, $\xi\sim \mathcal{N}(0, \mathcal{F}^{-1}\,\mathcal{C})$, $t \sim \mathcal{U}(0, T)$
     \State Extract start and goal configurations $(q_{\mathrm s}, q_{\mathrm g})$ from $x_0$
    \State Zero-End Projection $\tilde{x}_0(\tau) = x_0(\tau) - (1-\tau)q_{\mathrm{s}} - \tau q_{\mathrm{g}}$
    \State Forward diffusion process in coefficient space $\tilde{X}^k_{t} = \sqrt{\alpha_k(t)} \, \tilde{X}_{0}^k+\sqrt{1-\alpha_k(t)}\,\xi^k$
    \State Loss $\mathcal{L}(\theta)=\left\|\xi-\xi_\theta\!\left(\tilde{x}_t,t, q_\mathrm{s}, q_\mathrm{g} \right)\right\|_{2}^{2}$
    \State Update $\theta\leftarrow \theta-\eta \nabla_\theta \mathcal{L}(\theta)$
\EndWhile
\end{algorithmic}
\end{algorithm}

\begin{algorithm}[ht]
\caption{Guided Diffusion Inference}
\label{alg:infer}
\begin{algorithmic}[1]
\Require Pre-trained diffusion model $\xi_\theta$, start/goal states $(q_{\mathrm s}, q_{\mathrm g})$, covariance operator $\mathcal{F}^{-1}\mathcal{C}$, spectral schedule $\alpha_k(t)$, Sampling sequence $T=t_1>t_2>\cdots>t_M=0$
\Ensure Planned trajectory $x_0$

\State Sample initial noise $\tilde{x}_T \sim \mathcal{N}(\mathbf{0},\mathcal F ^{-1}\mathcal{C})$
\For{$i=1,2,\dots,M-1$}
    \State $t \gets t_i,\  s \gets t_{i+1}$
    \State Estimate denoised coefficients $\hat{\tilde X}_0^k \gets \frac{\tilde X_t^k-\sqrt{1-\alpha_k(t)}\,\xi_\theta^k}{\sqrt{\alpha_k(t)}}$
    \State Compute DDIM update
    $\tilde \mu_{s}^k
    = \sqrt{\alpha_k(s)}\,\hat{\tilde X}_0^k
    + \sqrt{1-\alpha_k(s)}\sqrt{1-\eta^2}\,\hat{\xi}_\theta^k$
    \State Recover trajectory $\hat x_0(\tau) = \hat{\tilde x}_0(\tau) + (1-\tau)q_{\mathrm{s}} + \tau q_{\mathrm{g}}$
    \State Compute guidance direction $\boldsymbol{g} \gets -\nabla_{\tilde \mu_s}\mathcal{J}(\hat x_0)$
    \State Apply classifier guidance
    \begin{equation}
        \tilde x_{s} \sim \mathcal{N}\!\left(\tilde \mu_{s}+\mathcal{F}^{-1}\mathcal{C} \,\boldsymbol{g},\, \eta^2\mathcal{F}^{-1}\mathcal{C}\right) \nonumber
    \end{equation}
\EndFor
\State Recover the trajectory $x_0(\tau)\gets \tilde x_0(\tau)+(1-\tau)q_{\mathrm s}+\tau q_{\mathrm g}$
\end{algorithmic}
\end{algorithm}

\begin{figure*}[htb]
    \centering
    \begin{subfigure}{0.23\linewidth}
    \centering
    \includegraphics[width=\textwidth]{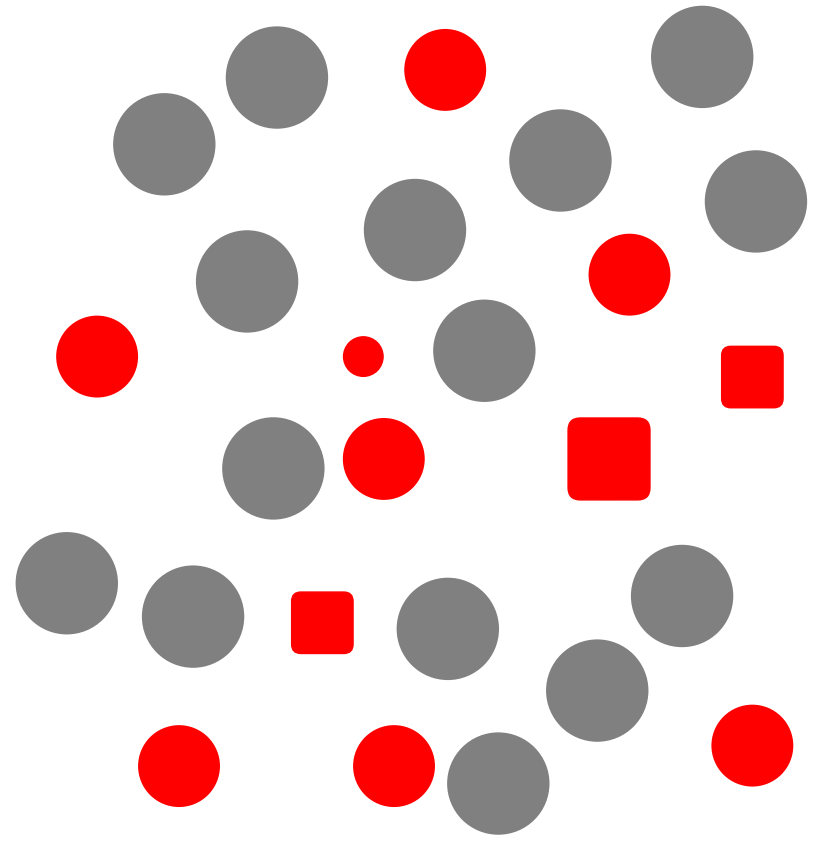}
    \caption{Simple2D}
    \end{subfigure}
    \begin{subfigure}{0.23\linewidth}
    \centering
    \includegraphics[width=\textwidth]{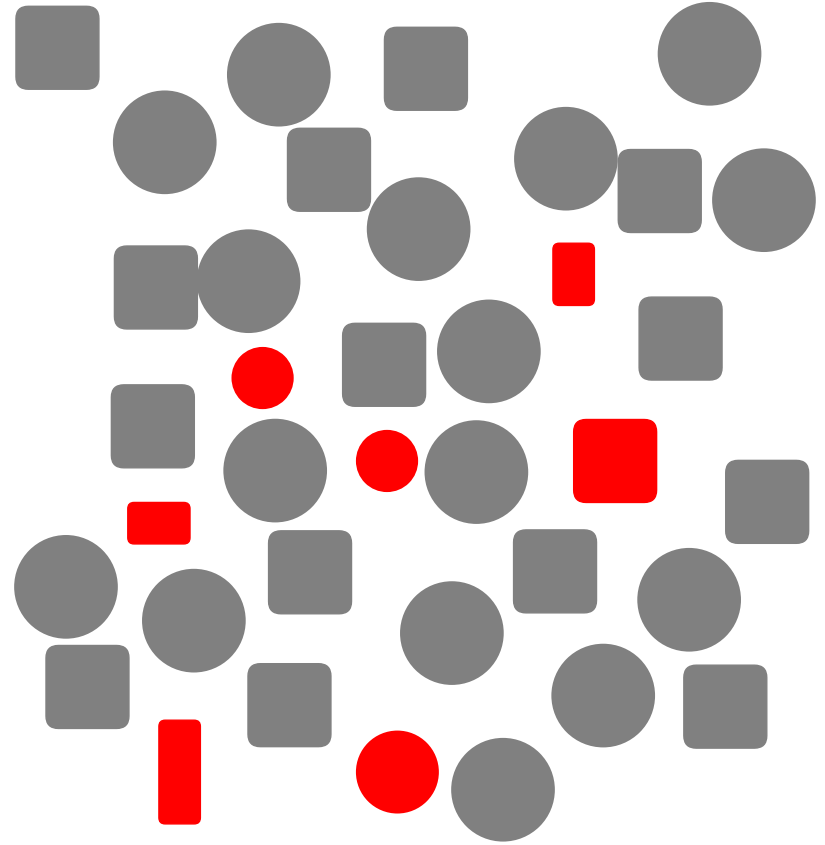}
    \caption{Dense2D}
    \end{subfigure}
    \begin{subfigure}{0.23\linewidth}
    \centering
    \includegraphics[width=\textwidth]{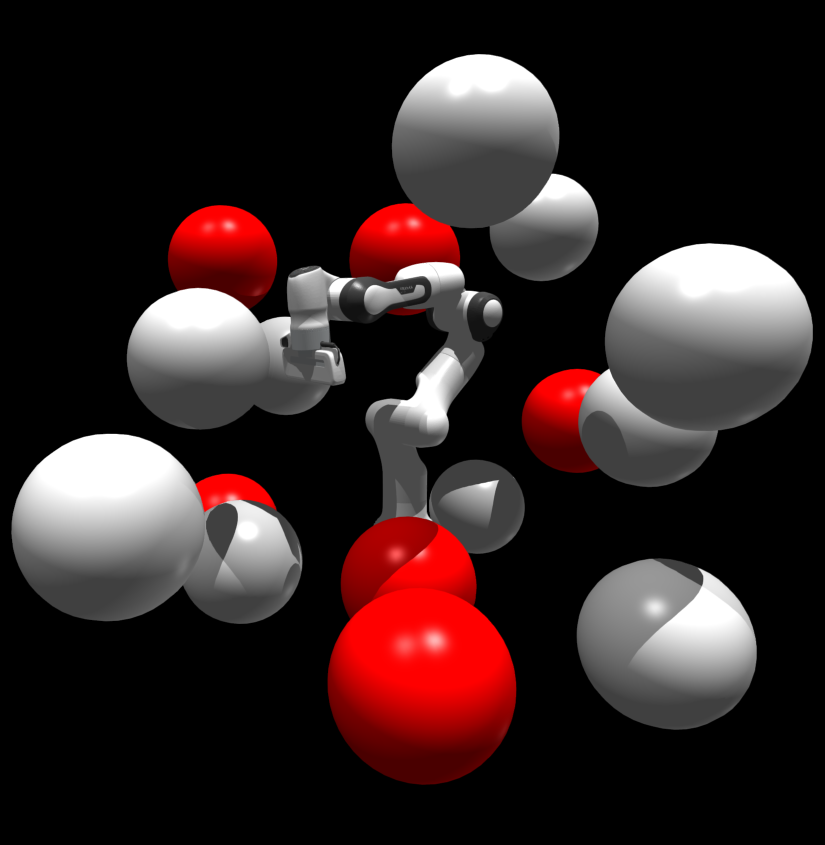}
    \caption{Spheres3D}
    \end{subfigure}
    \begin{subfigure}{0.23\linewidth}
    \centering
    \includegraphics[width=\textwidth]{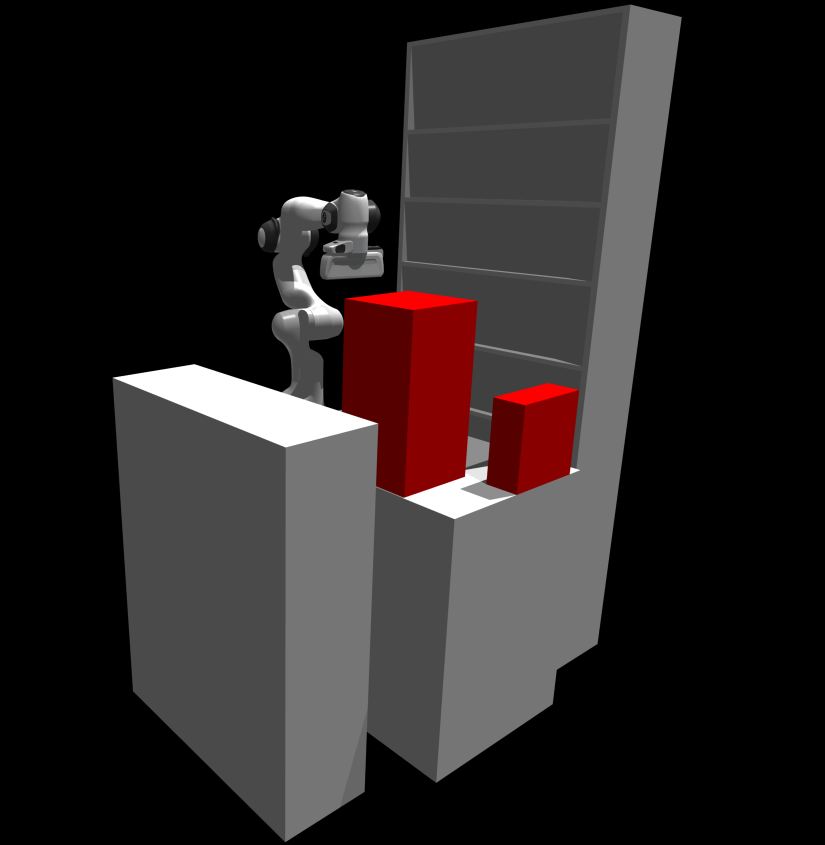}
    \caption{Warehouse}
    \end{subfigure}
    \caption{Environments used in the motion planning experiments. Gray obstacles are present during training, while red obstacles are introduced only at inference time to evaluate generalization to unseen obstacles.}
    \label{fig:envs}
\end{figure*}

\paragraph{\textbf{Boundary-Compatible DST-FNO}}
The zero-end projection imposes a boundary structure that should be preserved when the model is evaluated at resolutions different from the training resolution.
However, standard FNO (Sec~\ref{sec:fno}) relies on the FFT, whose periodic basis does not encode the zero-boundary condition. As a result, although the periodic basis fits the training-resolution data well, it may produce endpoint artifacts and degrade cross-resolution generalization.

To provide a boundary-compatible spectral representation, we replace the FFT with the discrete sine transform (DST), whose sine basis $\phi_k(s) = \sin(k\pi s)$ are naturally compatible with the endpoint structure. The Fourier layer becomes:
\begin{equation}
    (\mathcal{N} u)(\tau) = \sigma\left(Wu(\tau) + \mathcal{S}^{-1}(R \cdot \mathcal{S}u)(\tau)\right)
    \label{eq:dst-fno-layer}
\end{equation}
where $\mathcal{S}$ and $\mathcal{S}^{-1}$ denote the DST and inverse DST, and $R \in \mathbb{R}^{K \times c \times c}$ is a real-valued spectral kernel. For intermediate features with nonzero endpoints, we apply zero-end projection~\eqref{eq:zero-end-projection} before $\mathcal{S}$ and reconstruct afterward. 

We refer to this modified architecture as \emph{DST-FNO}, a minimal yet principled modification that provides the correct boundary inductive bias for cross-resolution generalization.

\subsection{Inference via Guided Sampling}
To generate collision-free trajectories, we sample from the learned prior $p(\tilde{x})$ and incorporate guidance during the reverse process. Since the forward process~\eqref{eq:closed-form-xi} admits a closed-form marginal for each spectral mode, we perform DDIM~\cite{song2021denoising} updates mode-wise in the spectral domain. It is also possible to use other more advanced inference algorithms \cite{zhang2022fast,zhang2022gddim}.

Given the noisy coefficient $\tilde{X}_t^k$ and the noise prediction $\xi_\theta^k(\tilde{x}_t,t,q_\mathrm{s}, q_\mathrm{g})$, we first estimate the clean coefficient:
\begin{equation}
    \hat{\tilde{X}}_0^k
    = \frac{1}{\sqrt{\alpha_k(t)}}\left(\tilde X_t^k - \sqrt{1-\alpha_k(t)}\,\xi^k_\theta\right)
    \label{eq:tweedie_mode}
\end{equation}
where $\xi_\theta$ is our trained DST-FNO. DDIM then transports $\tilde X_t^k$ to the next step $s<t$:
\begin{equation}
    \tilde X_{s}^k
    = \sqrt{\alpha_k(s)}\,\hat{\tilde X}_0^k
    + \sqrt{1-\alpha_k(s)}\left(\sqrt{1-\eta^2}\,\xi_\theta^k + \eta\, z^k\right)
    \label{eq:ddim_mode}
\end{equation}
where $z^k\sim\mathcal{N}(0,\mu_k^2/\lambda_k)$ and $\eta \in [0,1]$ controls the stochasticity.

The DDIM sampler generates trajectories from the learned prior but does not explicitly enforce collision avoidance. To incorporate obstacle constraints, we introduce a Boltzmann-form likelihood $p(\mathcal{O}\mid x)\propto \exp(-C(x))$, 
where $C(x)$ is a differentiable collision cost, and adopt a classifier guidance style update to incorporate it during sampling. A common practice is to evaluate this likelihood directly on the noisy sample $x_t$~\cite{janner2022diffuser,carvalho2023mpd}, but collision avoidance should be evaluated with respect to the final trajectory $x_0$. Formally, this likelihood requires marginalization over the unknown clean trajectory:
\begin{equation}
    p(\mathcal{O}\mid x_t) = \int p\left(\mathcal{O} \mid x_0\right)\, p\left(x_0\mid x_t\right) \,\mathrm{d}x_0
\end{equation}
which is intractable in practice. Following~\cite{chung2023diffusion, song2023loss}, we approximate it using the current clean prediction $\hat{x}_0$:
\begin{equation}
    p(\mathcal{O}\mid x_t) \simeq p(\mathcal{O}\mid\hat{x}_0) \propto \exp (-C(\hat{x}_0))
\end{equation}

This avoids computing collision gradients on highly corrupted samples and better aligns the guidance with the final denoised trajectory. 

During each reverse step, we first perform an unguided DDIM update using~\eqref{eq:ddim_mode} to obtain an intermediate prediction $\tilde{\mu}_s$, at which the guidance direction is evaluated:
\begin{equation}
    \boldsymbol{g}:=\left.\nabla_{x_{t}} \log p\left(\mathcal{O} \mid \hat{x}_{0}\right)\right|_{x_{t}=\tilde \mu_{s}}
\end{equation}

We then obtain the guided iterate through a covariance-preconditioned update~\cite{dhariwal2021diffusion}:
\begin{equation}
    \tilde x_s
    =
    \tilde \mu_s + \Sigma_t \boldsymbol{g},
    \label{eq:guided-ddim-update}
\end{equation}
where $\Sigma_t$ denotes the covariance preconditioner.

\section{Experiments}
\label{sec:experiment}

\subsection{Experimental Setup}
\paragraph{Environments and Datasets}
We evaluate FSD-MP across four environments covering 2D point robot and 3D manipulator planning, following the setup of MPD~\cite{carvalho2023mpd}. As shown in Fig.~\ref{fig:envs}, \textbf{Simple2D} and \textbf{Dense2D} involve a point robot navigating in 2D workspaces with sparse and dense obstacles, respectively. \textbf{Spheres3D} and \textbf{Warehouse} evaluate a 7-DoF Franka manipulator planning in 3D workspaces, where Spheres3D contains spherical obstacles, while Warehouse offers a more realistic environment with tables and a shelf. To test the generalization capability to unseen obstacles, each environment is further extended with additional obstacles not present during training.

For each environment, we generate an expert dataset using the same pipeline as MPD. We first use RRTConnect~\cite{kuffner2000rrt} to obtain initial feasible trajectories, then refine them with GPMP~\cite{mukadam2018continuous} to produce smooth demonstrations for diffusion training.

\paragraph{Baselines}
We compare FSD-MP against four baselines. \textbf{MPD}~\cite{carvalho2023mpd} is a diffusion-based planner trained and evaluated at a fixed resolution. 
\textbf{MPD-Spline}~\cite{carvalho2025mpd} fits training trajectories with B-spline curves and trains a diffusion model over the resulting low-dimensional control points, yielding a continuous trajectory representation.
\textbf{GPMP}~\cite{mukadam2018continuous} represents trajectories as continuous-time Gaussian processes and optimizes them via factor graph inference.
Additionally, we evaluate \textbf{MPD-Retrain}, which retrains MPD from scratch at each target resolution, to examine whether a single FSD-MP model can match resolution-specific retraining.

\paragraph{Metrics}
We evaluate the performance through two average metrics. Success rate measures the percentage of start-goal pairs for which at least one generated trajectory is collision-free. Valid rate measures the percentage of generated trajectories that are collision-free. Each method is evaluated over 120 randomly sampled start--goal pairs per environment, generating 100 trajectory samples per pair.

\subsection{Planning Results}
Tab.~\ref{tab:benchmarking} compares FSD-MP with baselines in both training and extra-obstacle environments (see Fig.~\ref{fig:envs}). In the training environments, diffusion-based methods are evaluated without guidance to assess the quality of the learned trajectory distribution. In the extra-obstacle settings, we introduce additional obstacles during inference and apply guidance to adapt to these unseen obstacles. GPMP is optimized directly in the corresponding environment for each setting.

At the training resolution, FSD-MP achieves competitive or higher success and valid rates compared with the baselines across both 2D point robot and 3D Franka manipulator environments. These results indicate that FSD-MP can generate high-quality samples and generalize across diverse planning problems, including high-dimensional manipulation tasks.

\begin{figure}[htb]
    \centering
    \begin{subfigure}{0.49\linewidth}
    \centering
    \includegraphics[width=\textwidth]{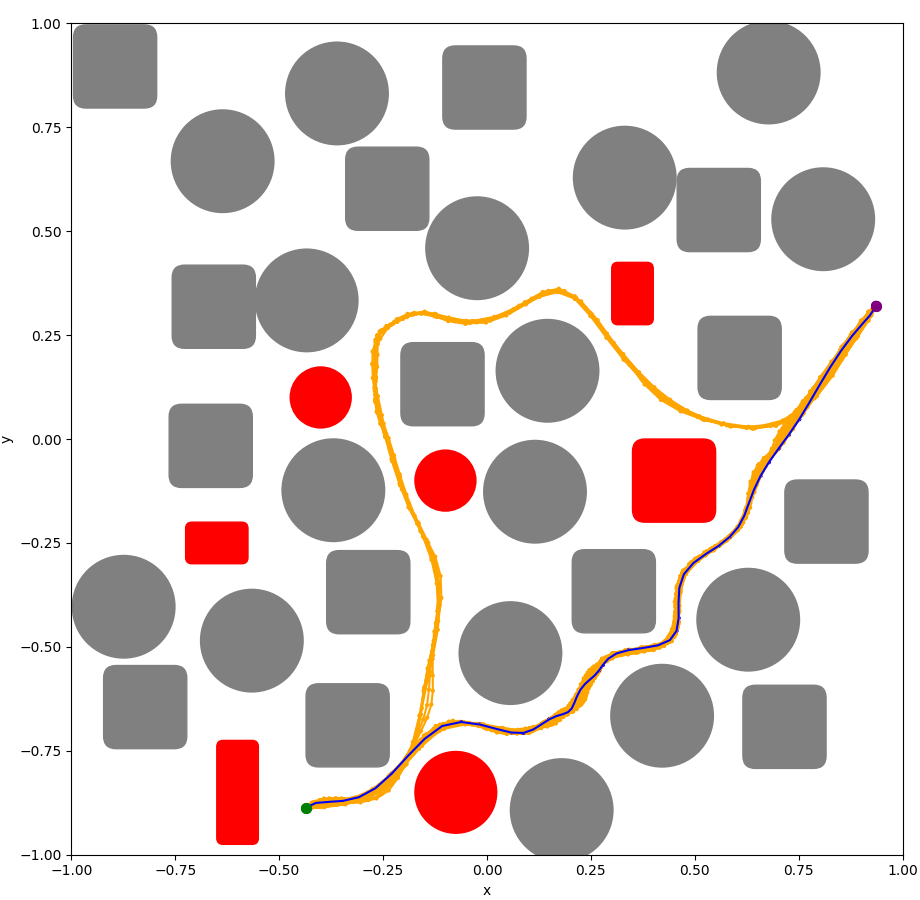}
    \caption{$N=64$, Context 1}
    \end{subfigure}
    \begin{subfigure}{0.49\linewidth}
    \centering
    \includegraphics[width=\textwidth]{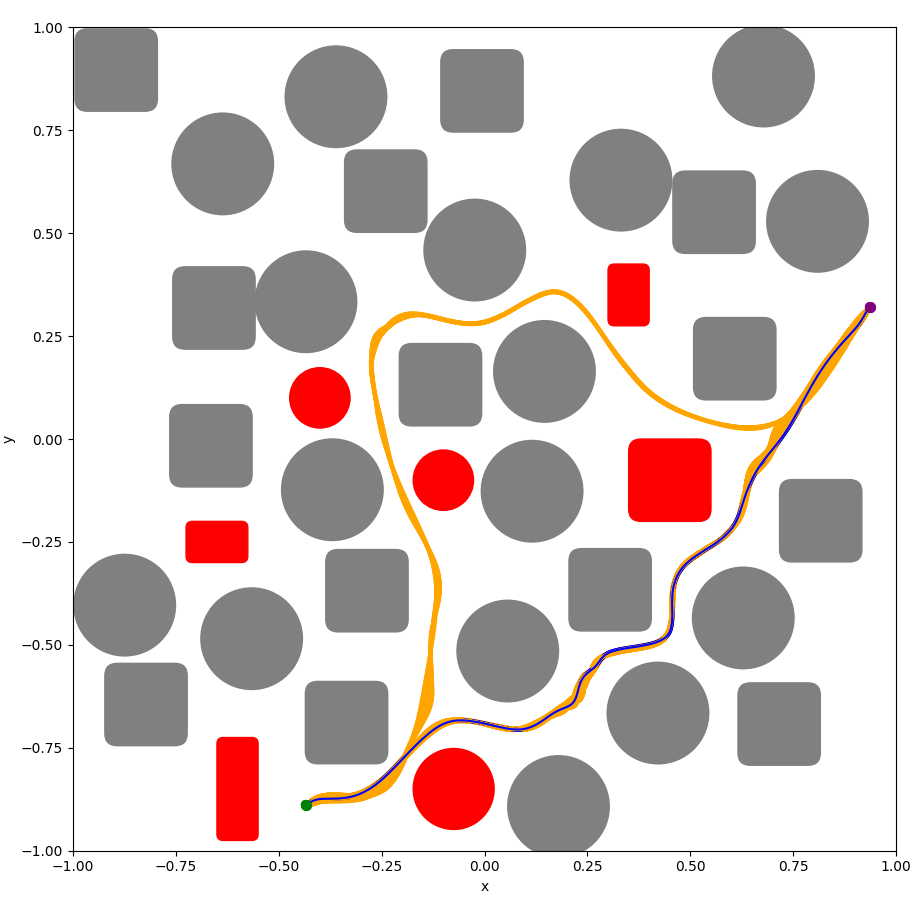}
    \caption{$N=1024$, Context 1}
    \end{subfigure}
    \begin{subfigure}{0.49\linewidth}
    \centering
    \includegraphics[width=\textwidth]{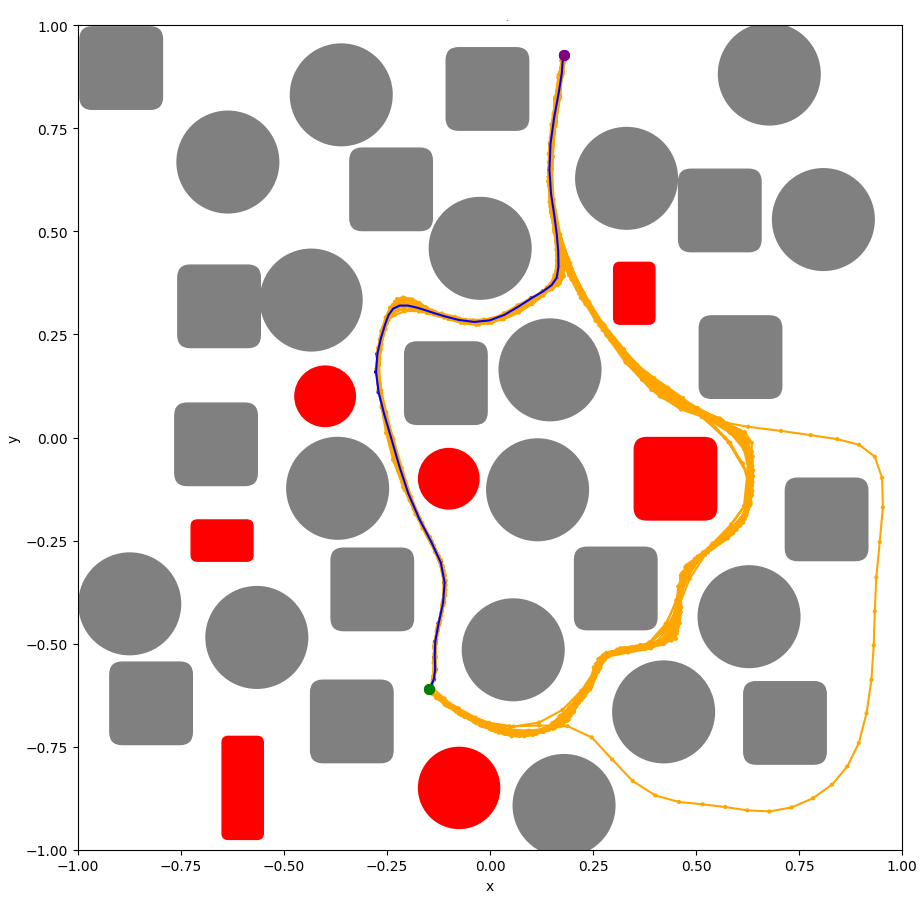}
    \caption{$N=50$, Context 2}
    \end{subfigure}
    \begin{subfigure}{0.49\linewidth}
    \centering
    \includegraphics[width=\textwidth]{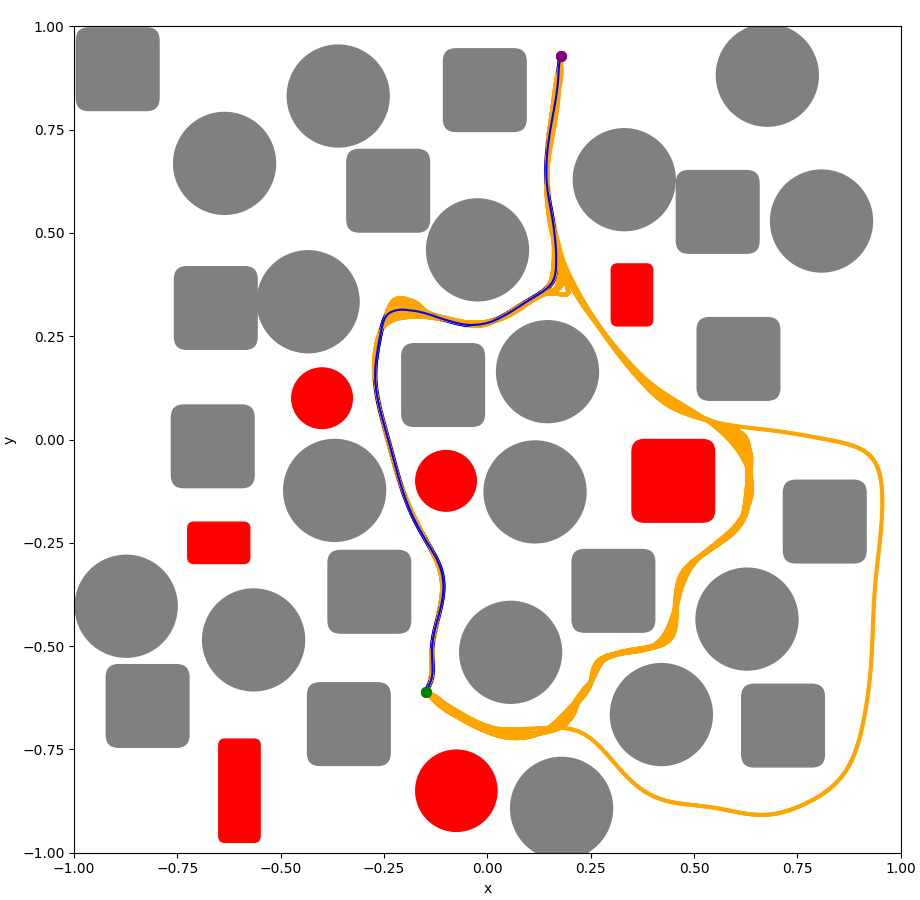}
    \caption{$N=2000$, Context 2}
    \end{subfigure}
    \caption{FSD-MP planning results in Dense2D with extra obstacles. The model is trained only at resolution \(N=64\) and evaluated at multiple resolutions without retraining.}
    \label{fig:dense_cross_resolution}
\end{figure}

To evaluate cross-resolution generalization, we train the model only at resolution $N=64$ and perform inference at unseen resolutions without retraining. As shown in Tab.~\ref{tab:benchmarking}, FSD-MP maintains nearly identical performance across $N=64$ and $N=1024$, demonstrating zero-shot cross-resolution inference. 
Fig.~\ref{fig:dense_cross_resolution} further visualizes Dense2D results with extra obstacles at \(N=50\), \(N=64\), \(N=1024\), and \(N=2000\), yielding nearly identical trajectories across resolutions.

\begin{figure*}[ht]
    \centering
    \begin{subfigure}{0.23\linewidth}
    \centering
    \includegraphics[width=\textwidth]{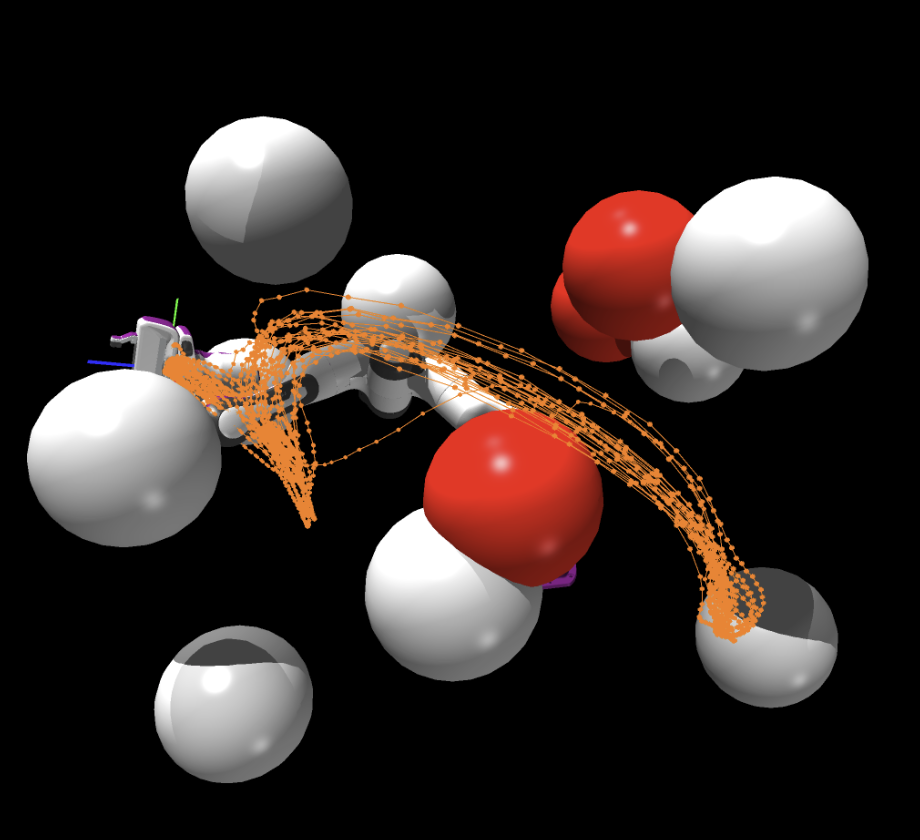}
    \caption{Spheres3D, Context 1}
    \end{subfigure}
    \begin{subfigure}{0.23\linewidth}
    \centering
    \includegraphics[width=\textwidth]{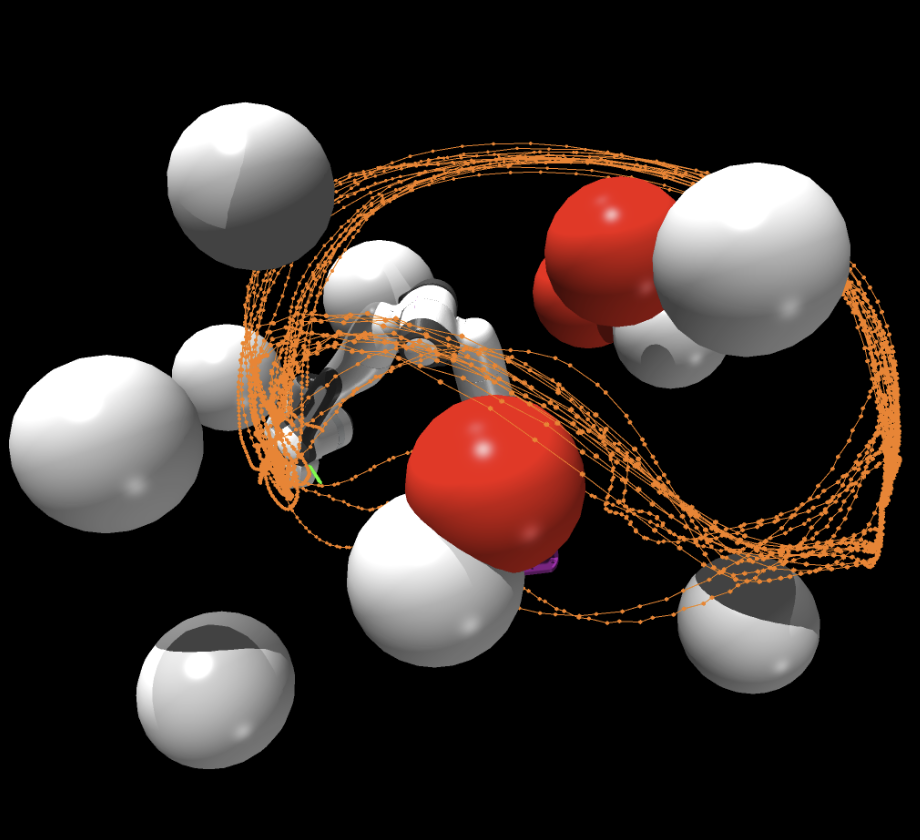}
    \caption{Spheres3D, Context 2}
    \end{subfigure}
    \begin{subfigure}{0.23\linewidth}
    \centering
    \includegraphics[width=\textwidth]{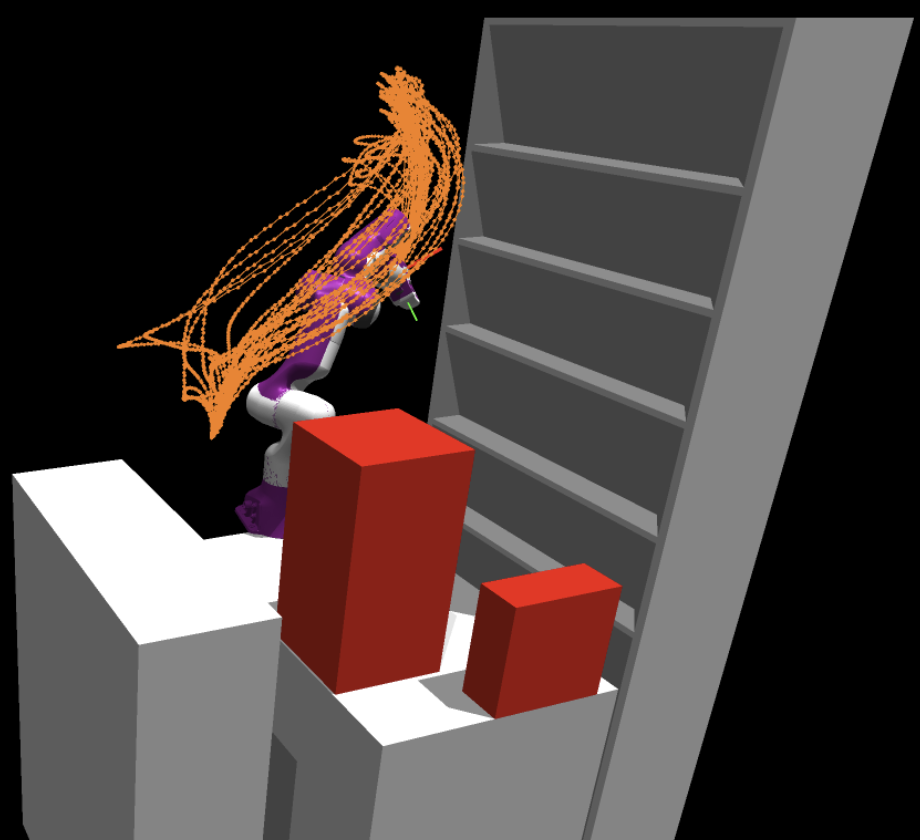}
    \caption{Warehouse, Context 1}
    \end{subfigure}
     \begin{subfigure}{0.23\linewidth}
    \centering
    \includegraphics[width=\textwidth]{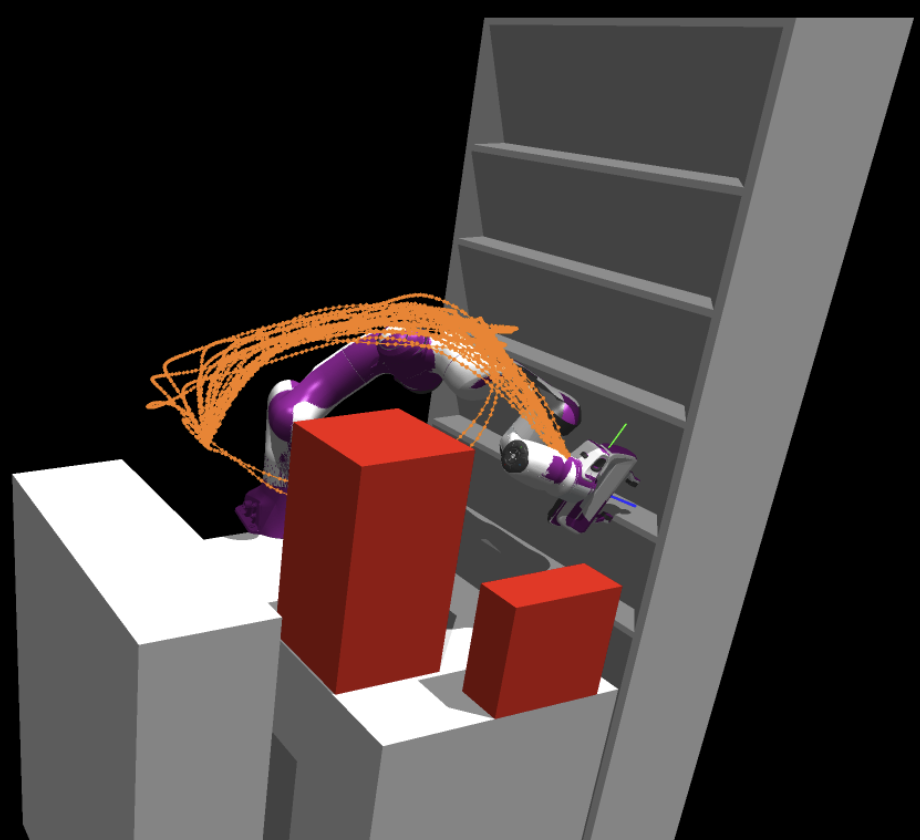}
    \caption{Warehouse, Context 2}
    \end{subfigure}
    \caption{Planning results of FSD-MP in 3D Franka arm environments. Orange lines are end-effector trajectories.}
    \label{fig:3d_results}
\end{figure*}

Fig.~\ref{fig:3d_results} visualizes representative planning results of FSD-MP in the 3D Franka environments. Across different start-goal contexts in both Spheres3D and Warehouse, the generated trajectories remain smooth and avoid the additional obstacles.

\begin{table*}[htb]
\centering
\caption{Benchmarking in different resolutions and environments}
\label{tab:benchmarking}
\begin{tabular}{llcccccccc}
\toprule
& & \multicolumn{4}{c}{Training Environment (Prior Only)} & \multicolumn{4}{c}{Extra Obstacles (Guided)}\\
\cmidrule(lr){3-6} \cmidrule(lr){7-10} 
& & \multicolumn{2}{c} {\textbf{$N=64$}} & \multicolumn{2}{c} {\textbf{$N=1024$}}& \multicolumn{2}{c} {\textbf{$N=64$}} & \multicolumn{2}{c} {\textbf{$N=1024$}}\\
\cmidrule(lr){3-4} \cmidrule(lr){5-6} \cmidrule(lr){7-8} \cmidrule(lr){9-10}
\textbf{Env} & \textbf{Method} & Succ. (\%) & Valid (\%)  & Succ. (\%) & Valid (\%) & Succ. (\%) & Valid (\%) & Succ. (\%) & Valid (\%)\\
\midrule
\multirow{5}{*}{\rotatebox[origin=c]{90}{\textsc{Simple2D}}}
& GPMP & 100 & 98.6 & 100 & 99.8 & 100 & 98.0 & 99.2 & 96.9 \\
& MPD & 100 & 99.7 & -- & -- & 100 & 91.6 & -- & --\\
& MPD-Retrain & -- & -- & 100 & 99.8 & -- & -- & 100 & 94.4\\
& MPD-Spline & 100 & 94.3 & 100 & 92.8 & 100 & 76.3 & 100 & 75.6 \\
& FSD-MP (Ours) & 100 & 99.9 & 100 & 99.8 & 100 & 97.7 & 100 & 98.4 \\
\midrule
\multirow{5}{*}{\rotatebox[origin=c]{90}{\textsc{Dense2D}}}
& GPMP & 98.3 & 58.3 & 75.8 & 50.7 & 91.7 & 50.2 & 35.8 & 24.5 \\
& MPD & 100 & 99.7 & -- & -- & 89.2 & 54.3 & -- & --\\
& MPD-Retrain & -- & -- & 100 & 99.8 & -- & -- & 90 & 63.7  \\
& MPD-Spline & 91.7 & 71.5 & 85 & 54.1 & 83.3 & 31.1 & 79.2 & 28.5\\
& FSD-MP (Ours) & 100 & 99.8 & 100 & 99.9 & 91.7 & 62.6 & 91.7 & 66.4  \\
\midrule
\multirow{5}{*}{\rotatebox[origin=c]{90}{\textsc{Spheres3D}}}
& GPMP & 95.8 & 79.6 & 92.5 & 89.3 & 91.7 & 76.4 & 90.0 & 80.2 \\
& MPD & 95 & 71.7 & -- & -- & 89.2 & 62.0 & -- & -- \\
& MPD-Retrain & -- & -- & 96.7 & 73.3 & -- & -- & 92.5 & 72.8 \\
& MPD-Spline & 80.8 & 27.2 & 80 & 26.1 & 86.7 & 42.3 & 83.3 & 39.6\\
& FSD-MP (Ours) & 100 & 83.8 & 100 & 83.4 & 96.7 & 75.5 & 96.7 & 75.5 \\
\midrule
\multirow{5}{*}{\rotatebox[origin=c]{90}{\textsc{Warehouse}}}
& GPMP & 99.2 & 92.1 & 93.3 & 86.3 & 99.2 & 95.0 & 91.7 & 85.4 \\
& MPD & 94.2 & 88.5 & -- & -- & 99.2 & 98.7 & -- & -- \\
& MPD-Retrain & -- & -- & 95.0 & 88.8 & -- & -- & 99.2 & 99.1 \\
& MPD-Spline  & 92.5 & 75.9 & 91.7 & 74.6 & 96.7 & 93.2 & 96.7 & 94.5\\
& FSD-MP (Ours) & 95.8 & 90.3 & 95.8 & 90.2 & 100 & 100 & 100 & 100 \\
\bottomrule
\end{tabular}
\end{table*}

\subsection{Ablation Studies}
We conduct ablation studies on Dense2D and Warehouse to validate the key design components of FSD-MP. We evaluate three variants: \textbf{FNO}, which replaces the DST-based spectral layers with standard FFT layers; \textbf{White Noise}, which replaces the Mat\'ern-type covariance with an identity covariance; and \textbf{UNet}, which replaces the DST-FNO backbone with a UNet of comparable parameter count. All models are trained at $N=64$ and evaluated across multiple resolutions without retraining. Table~\ref{tab:ablation} reports the collision-free rate.

FSD-MP maintains nearly constant performance across resolutions in both environments, whereas the ablated variants degrade under cross-resolution inference. The standard FNO performs well near $N=64$ but drops substantially at high resolutions, showing the importance of the boundary-compatible DST basis. White noise consistently underperforms FSD-MP across all resolutions, indicating the benefit of well-defined function-space perturbations. The UNet variant performs well only at the training resolution and fails to generalize across resolutions, confirming the need for a resolution-agnostic neural operator.

\begin{table*}[htb]
\centering
\caption{Ablation on Architecture and Diffusion Noise: Valid Rate (\%) Across Resolution $N$}
\label{tab:ablation}
\begin{tabular}{lcccccccccc}
\toprule
& \multicolumn{5}{c}{Dense2D} & \multicolumn{5}{c}{Warehouse}\\
\cmidrule(lr){2-6} \cmidrule(lr){7-11}
Method & $N$=32 & $N$=64 & $N$=256 & $N$=512 & $N$=1024 & $N$=32 & $N$=64 & $N$=256 & $N$=512 & $N$=1024 \\
\midrule
FSD-MP & \textbf{99.5} & \textbf{99.8} & \textbf{99.8} & \textbf{99.8} & \textbf{99.8} & \textbf{90.8} & 90.3 & \textbf{90.1} & \textbf{90.2} & \textbf{90.2}\\
FNO & 95 & 99.6 & 88.7 & 73.4 & 63.5 & 90.1 & \textbf{90.5} & 84.8 & 80.6 & 77.2 \\
White Noise & 94.7 & 96.0 & 96.7 & 96.8 & 97.0 & 85.8 & 86.1 & 86.4 & 86.6 & 86.8 \\
UNet & 0.3 & 97.3 & 0.1 & 0 & 0 & 64.2 & 87.2 & 51.6 & 38.8 & 26.4 \\ 
\bottomrule
\end{tabular}
\end{table*}

\section{Conclusion}
\label{sec:conclusion}
We presented Function-Space Diffusion for Motion Planning (FSD-MP), a diffusion-based motion planner that operates in infinite-dimensional function space. By combining function-space diffusion with a boundary-compatible DST-FNO, FSD-MP achieves zero-shot generalization across arbitrary resolutions without retraining. Experiments on 2D point robot and 3D Franka manipulator environments demonstrate competitive planning performance at the training resolution and nearly identical performance under cross-resolution inference. Future work will explore extension to higher-dimensional configuration spaces, coarse-to-fine inference, and adaptive-resolution generation with local refinement near dense obstacles or narrow passages.

\appendix

\subsection{Objective for FSD-MP}
\label{sec:objective}

We derive the training objective from the conditional distribution $p(x_t \mid x_0)$ in the function space. 
From the closed-form marginal~\eqref{eq:closed-form-xi}, the coefficients $\{X_t^k\}$ are conditionally independent given $x_0$, with
\begin{equation}
    p(x_t \mid x_0)
    = \prod_{k=1}^{\infty} p(X_t^k \mid X_0^k)
\end{equation}
where each component follows a Gaussian distribution.

The score in the function space is defined as
$\nabla_{x_t} \log p(x_t \mid x_0)$, which admits the expansion in the basis $\{\phi_k\}$:
\begin{equation}
    \nabla_{x_t} \log p(x_t \mid x_0)
    =
    \sum_{k=1}^{\infty}
    \partial_{X_t^k} \log p(X_t^k \mid X_0^k)\,\phi_k
\end{equation}

We consider score matching in the function space, which minimizes
\begin{equation}
    \mathbb{E}
    \bigl[
        \| s_\theta(x_t,t) - \nabla_{x_t} \log p(x_t \mid x_0) \|_{\mathcal H}^2
    \bigr].
\end{equation}

Using the orthonormal basis $\{\phi_k\}$, this objective decomposes into
\begin{equation}
    \mathbb{E}
    \left[
        \sum_{k=1}^{\infty}
        \left|
            s_\theta^k(x_t,t)
            -
            \partial_{X_t^k} \log p(X_t^k \mid X_0^k)
        \right|^2
    \right].
\end{equation}

For each mode, the conditional distribution is Gaussian, and its score is linearly related to the noise $\xi^k$. Up to a mode-dependent scalar, predicting the score is equivalent to predicting the noise $\xi^k$. This leads to the objective
\begin{equation}
    \mathcal L(\theta) = \mathbb E_{t,x_0,\xi}
    \left[
        \sum_{k=1}^{\infty}
        \left|
            \xi_\theta^k(x_t,t)-\xi^k
        \right|^2
    \right].
\end{equation}

Finally, Parseval's identity $\sum_{k=1}^{\infty} \left|\xi_\theta^k-\xi^k\right|^2 = \|\xi_\theta-\xi\|_{\mathcal H}^2$ yields
\begin{equation}
    \mathcal L(\theta)
    =
    \mathbb E_{t,x_0,\xi}
    \bigl[
        \|\xi_\theta(x_t,t)-\xi\|_{\mathcal H}^2
    \bigr]
\end{equation}
which recovers the objective in~\eqref{eq:xi-loss-time-domain}.

\subsection{Trace-Class Gaussian Noise in Function Space}
\label{sec:noise}
A Gaussian measure $\mathcal{N}(0, \mathcal{C})$ on $\mathcal{H}$ is the infinite-dimensional analogue of a multivariate Gaussian noise, characterized by a covariance operator $\mathcal{C}: \mathcal{H} \to \mathcal{H}$. It is supported on $\mathcal{H}$ only if $\mathcal{C}$ is trace-class, self-adjoint, and positive, ensuring finite expected energy $\mathbb{E}\left[\|x\|^2_\mathcal{H}\right] = \operatorname{tr}(\mathcal{C}) < \infty$. 
The identity operator fails this requirement since $\operatorname{tr}(\mathcal{I}) = \sum_{k=1}^{\infty} 1 = \infty$, making white noise $\mathcal{N}(0, \mathcal{I})$ ill-defined in $\mathcal{H}$.

The Mat\'ern-type operator $\mathcal{C} = \sigma^2(-\Delta + \kappa^2 \mathcal{I})^{-\alpha}$ satisfies all three conditions when $\alpha$ is sufficiently large. Its covariance eigenvalues decay as $\lambda_k\sim k^{-2\alpha/D}$, giving $\operatorname{tr}(\mathcal{C}) = \sum_k \lambda_k < \infty$ when $\alpha > D/2$, where $D$ is the input dimension of the function. For time-parameterized trajectories, $D=1$ and the condition reduces to $\alpha > 1/2$.

\bibliographystyle{Bibliography/IEEETrans}
\bibliography{references}

\end{document}